\documentclass[10pt,twocolumn,letterpaper]{article}

\usepackage{iccv}     
\usepackage{times}
\usepackage{epsfig}

\usepackage{graphicx}
\usepackage{amsmath}
\usepackage{amssymb}
\usepackage{booktabs}
\usepackage{tikz}
\usepackage{multirow}
\usepackage{makecell}
\def \names {STPD }

\usepackage[pagebackref=true,breaklinks=true,letterpaper=true,colorlinks,bookmarks=false]{hyperref}

\usepackage[capitalize]{cleveref}
\crefname{section}{Sec.}{Secs.}
\Crefname{section}{Section}{Sections}
\Crefname{table}{Table}{Tables}
\crefname{table}{Tab.}{Tabs.}
\makeatletter
\def\thanksno#1{\protected@xdef\@thanks{\@thanks
        \protect\footnotetext{#1}}}
\makeatother

\def\iccvPaperID{5884} 

\iccvfinalcopy
\ificcvfinal\fi

\begin{document}

\title{Sketch and Text Guided Diffusion Model for Colored Point Cloud Generation}

\author{Zijie Wu$ ^{1}$\thanks{Work performed during visit at the University of Western Australia}
\quad Yaonan Wang$ ^{1}$ 
\quad Mingtao Feng$ ^{2}$\thanks{Correponding author.} 
\quad He Xie$ ^{1}$ 
\quad Ajmal Mian$ ^{3}$\\
\normalsize $ ^1 $Hunan University \quad
\normalsize $ ^2 $Xidian University \quad
\normalsize $ ^3 $The University of Western Australia\\
}
\maketitle

\vspace{-6mm}
\begin{abstract}
\vspace{-4mm}
%
Diffusion probabilistic models have achieved remarkable success in text guided image generation. However, generating 3D shapes is still challenging due to the lack of sufficient data containing 3D models along with their descriptions. Moreover, text based descriptions of 3D shapes are inherently ambiguous and lack details.  
In this paper, we propose a sketch and text guided probabilistic diffusion model for colored point cloud generation that conditions the denoising process jointly with a hand drawn sketch of the object and its textual description.
We incrementally diffuse the point coordinates and color values in a joint diffusion process to reach a Gaussian distribution. Colored point cloud generation thus amounts to learning the reverse diffusion process, conditioned by the sketch and text, to iteratively recover the desired shape and color. 
Specifically, to learn effective sketch-text embedding, our model adaptively aggregates the joint embedding of text prompt and the sketch based on a capsule attention network. Our model uses staged diffusion to generate the shape and then assign colors to different parts conditioned on the appearance prompt while preserving precise shapes from the first stage. This gives our model the flexibility to extend to multiple tasks, such as appearance re-editing and part segmentation. Experimental results demonstrate that our model outperforms recent state-of-the-art in point cloud generation.

\end{abstract}

\vspace{-6mm}
\section{Introduction}
\label{sec:intro}
\vspace{-2mm}
Denoising Diffusion Probabilistic Models (DDPM)~\cite{ho2020denoisingdiffusion1, sohl2015deepdiffusion2} can generate novel images and videos, from their text descriptions. Recent diffusion models \cite{pmlr-v139-ramesh21a, poole2022dreamfusion_text23d, Rombach_2022_CVPR, sanghi2022clip} contain billions of parameters and are trained on millions of text-image~\cite{feng2021free} pairs sourced from the Internet, including existing datasets like MS-COCO \cite{MSCOCO}. The availability of such large corpora of text-image pair datasets is, arguably, the major driving force behind the success of research in image and video generation. However, such datasets of text-shape pairs are almost nonexistent causing a major bottleneck for advancing research in this direction.

\begin{figure}[t!] 
		
		\begin{tikzpicture}[inner sep=0pt,outer sep=0pt]
		\centering
		\node[anchor=south west] (A) at (0in,0in)
		{\includegraphics[width=0.48\textwidth,clip=false]{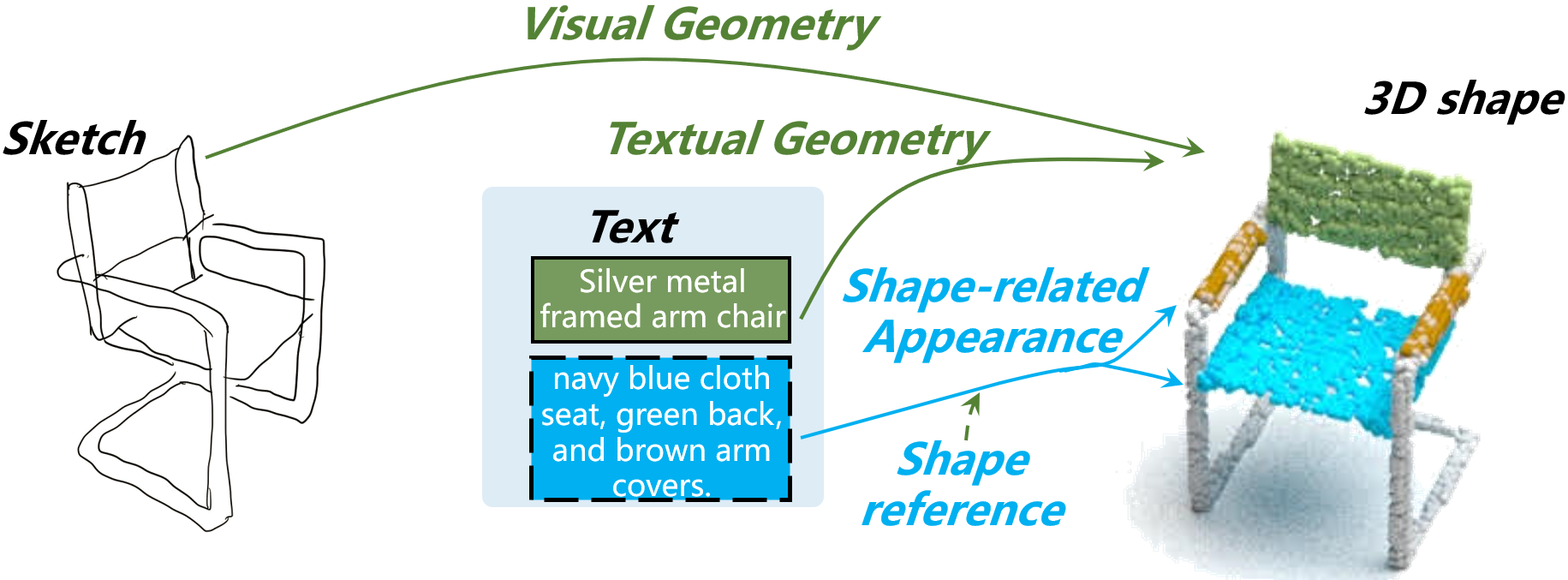}};
	
		\end{tikzpicture}
		
		\vspace{-2mm}
		\caption{\label{fig:sketchtext} Illustration of sketch and text guided 3D point cloud generation. The text and sketch complement each other.}
		\vspace{-7mm}
	\end{figure}
 
3D shape generation has a wide range of applications including data augmentation, virtual/augmented reality~\cite{feng2023exploring}, manufacturing, and reverse engineering. 
Pioneering works in diffusion based 3D shape generation include Luo~\textit{et al.}~\cite{diffusion1} and Point Voxel Diffusion (PVD)~\cite{zhou20213d}. Luo~\textit{et al.}~\cite{diffusion1} extend the DDPMs to 3D shape generation conditioned on
shape latent that follows a prior distribution parameterized via normalized flows~\cite{dinh2016density}. Their probabilistic generative model treats points as particles that, under heat, are diffused to a noise distribution. It then exploits the reverse diffusion process to learn the point distribution to be able to generate plausible point clouds. PVD~\cite{zhou20213d} combines the denoising diffusion model with the point-voxel 3D shape representation. It performs the denoising steps by optimizing a variational lower bound to the conditional likelihood function to produce 3D shapes or multiple 3D completions from a partial observation.

The problem of text guided 3D model or point cloud generation is slightly different from that of images in the sense that there is a higher emphasis, at least implicitly, on geometry rather than appearance which puts yet another constraint on the training datasets. Current methods \cite{diffusion1, zhou20213d} have mostly used the ShapeNet \cite{chang2015shapenet} dataset for text-shape learning combined with the ModelNet \cite{modelnet} datasets for 3D representation learning and validation. However, these datasets are relatively small and lack detailed textual descriptions. To work around the data paucity, CLIP-forge~\cite{sanghi2022clip} extends the CLIP model~\cite{radford2021learning} for 3D shape generation by aligning 3D shapes with text in the latent space. Although these text guided methods maintain diversity in the generated shapes, their common limitation is that the generated shapes are often not the desired ones, which is understandable given the shape ambiguity in the text.


In this paper, to reduce the shape ambiguity, we propose a sketch and text guided probabilistic diffusion (STPD) model for colored point cloud generation that conditions the denoising process with sketch inputs, in addition to textual descriptions. Sketches can be hand drawn and give much more geometric details compared to text. We argue that, combined with text, sketches are a viable option for conditioning the denoising of 3D shapes given their direct relevance to the problem at hand and their history of successful use for 3D object reconstruction~\cite{sketch2model}. 
However, cross-modality inputs (sketch and text)~\cite{feng2020relation} always suffer from large cross-modality discrepancies and intra-modality variations bringing in additional challenges of cross-modality fusion and data sparsity. Especially, sketches are inherently sparse with only a very small proportion of informative pixels, and contain much less information than natural images. 





The proposed STPD model can produce the geometry and appearance simultaneously and unambiguously from a sketch-text input, combining ease of access and completeness of description. 
It takes the sketch as the main shape description and shape information present in the text to complement the sketch input. Hence, shape information is extracted from the sketch as well as the text when there is shape information in the text. Appearance is extracted from the text input to give color to the generated point clouds as shown in Fig.~\ref{fig:sketchtext}. 
To summarize, our contributions are: 
\begin{enumerate}
\vspace{-1mm}
    \item We propose a staged probabilistic diffusion model for better control over the geometry and appearance to generate colored point clouds. Our model conditions the denoising process jointly with sketch and text inputs to generate 3D shapes. It can also be used for part-segmentation and appearance re-editing.
\vspace{-2mm}
    \item We propose an attention capsule based feature extraction module for encoding sketches, which are inherently sparse. Our method robustly gives more attention to the entities of the useful pixels and ignores the large number of meaningless capsules corresponding to the blank pixels in the sketch.
\vspace{-2mm}
    \item We present a sketch-text fusion network that efficiently abstracts the shape and color information from both sketch and language descriptions to guide the reverse 3D shape diffusion process.
  \vspace{-1mm}
  
\end{enumerate}

Extensive experiments on the ShapeNet dataset \cite{chang2015shapenet} and comparison to existing diffusion probabilistic model based generation as well as some classical shape reconstruction methods 
show that our model achieves state-of-the-art performance for colored point cloud generation. We also show the representation learning ability of our model by conducting 3D object classification experiments on the ModelNet40 \cite{modelnet} dataset and the application of our model to part segmentation on the ShapeNet-Parts dataset \cite{chang2015shapenet}.

\vspace{-1mm}
\section{Related Work}
\vspace{-2mm}

We first survey diffusion models for generating images. Next, we discuss 3D shape reconstruction methods followed by classical 3D shape generation methods. 
Finally, we discuss diffusion models for 3D shape generation.

\vspace{-1mm}
\subsection{Text-based 2D Image Generation}
\vspace{-1mm}

\noindent {\bf Generative Adversarial Network (GAN):} Early methods for text based image generation were based on GANs.  Xu~\textit{et al.}~\cite{xu2018attngan_text1} used attention mechanism for fine-grained text to image generation. TediGAN~\cite{Xia_2021_CVPRtext4} uses GAN inversion based on the pretrained StyleGAN~\cite{bermano2022state} for text to image generation. Zhang et al.\cite{zhang2017stackgantext1}\cite{zhang2018stackgan++text1} used multiple GAN architectures that stacked generators and discriminators to progressively generate multi-scale images. GANs are known to be difficult to train and have recently been surpassed by diffusion models~\cite{dhariwal2021diffusion_image}.

\noindent {\bf Probabilistic Diffusion Models:} 
Sohl-Dickstein et al.~\cite{sohl2015deepdiffusion2} proposed diffusion probabilistic models that are both tractable and flexible. 
Inspired by non-equilibrium thermodynamics, the diffusion process uses Markov chain~\cite{markov} to convert a simple Gaussian distribution to a target data distribution. 
Keeping the steps of the chain small enough, each step can be analytically evaluated~\cite{xiade}, and hence, the full chain can also be evaluated. At each step of the forward pass, a small perturbation is added to the data until it reaches the Gaussian distribution. Since the perturbation at each step is small, a model can be learned to estimate and remove the perturbation. It was not until 2020 that Ho et al.~\cite{ho2020denoisingdiffusion1} showed the real power of diffusion models with high quality image generation by training on a weighted variational bound and denoising score matching with Langevin dynamics. Dhariwal and Nichol~\cite{dhariwal2021diffusion_image} showed the superiority of diffusion models over GANs. However, directly optimizing in pixel space comes with a high computational cost. To reduce the computational requirements, \cite{Rombach_2022_CVPR} performs diffusion in the latent feature representation space. Large models such as Imagen \cite{Imagen} and DALL-E2~\cite{pmlr-v139-ramesh21a}, which uses the CLIP model~\cite{radford2021learning}, can generate diverse photo-realistic images. The success of diffusion models is largely attributed to fundamental research as well as the availability of huge datasets in the image domain. 

\vspace{-1mm}
\subsection{3D Reconstruction from Images}
\vspace{-1mm}
\noindent Reconstructing 3D shapes from images has been studied for decades. Single-view reconstruction techniques are popular, however, recovering 3D shapes from a single image is an ill-posed problem, as it requires a strong prior shape knowledge. Fan~\textit{et al.}~\cite{PSGN} proposed to directly regress point clouds from a single image. Wang~\textit{et al.}~\cite{Pix2mesh} deformed the ellipsoid mesh to form the 3D shape to learn the shape prior. Sketch2Mesh~\cite{Sketch2mesh} employed the silhouettes matching match between a given single sketch and the learned model to refine the 3D shape. TMNet~\cite{TMnet} uses triangular meshes and images to supervise the reconstruction of 3D shapes.

Multiview images contain more geometric information useful for 3D reconstruction. Pix2Vox++~\cite{xie2020pix2vox++} fuse the multiple coarse 3D volumes, which are generated from each individual view, to form the well-refined 3D shapes. Wang~\textit{et al.}~\cite{EvolT} include transformers to recover 3D information from multiviews.  Reconstruction based methods usually demand exact image-to-shape match and strong prior shape knowledge, whereas generative methods offer a balance between diversity and alignment with user specifications.

\vspace{-1mm}
\subsection{Sketch, Text and 3D Shape Generation}\vspace{-1mm}



\noindent {\bf Sketch to 3D Shape:} Sketches have been used as a sparse representation of natural images and 3D shapes. Sketches are quite illustrative, despite their abstract nature and simplicity. Sketch based 3D generation~\cite{olsen2009sketch_survey} 
can be coarsely categorized into evocative~\cite{wang2015sketch1} and constructive~\cite{miao2015symmsketch2} methods, which depend on retrieval from model collections or free-form modeling. 
Deep models learn the pattern of the sketch for 3D model generation.
For example, Zhang~\textit{et al.}~\cite{sketch2model} proposed a view-aware technique to gain priors from existing 3D models using a single freehand sketch. DeepSketchHair~\cite{sketch2model2tv} proposed a learning based system for intuitive 3D hairstyling from 2D sketches.
These methods have less dependence on a shape prior to the generation phase, but the tremendous requirement of shape-related data during training for pattern learning.

\noindent {\bf Text to 3D Shape:} Text to shape generation is a relatively new direction. Examples include Text2shape~\cite{chen2018text2shape} and CLIP-Forge~\cite{sanghi2022clip}. Text2shape~\cite{chen2018text2shape} generates colored 3D shapes using joint embedding of text and 3D shapes. CLIP-Forge~\cite{sanghi2022clip} generates 3D shapes from the pre-trained text-to-image CLIP~\cite{radford2021learning}, addressing the lack of text-to-shape priors data.  
Although previous studies have considered sketch and text as flexible methods for 3D shape generation, they only refine the model in an implicit pattern that takes the shape priors library into optimization while ignoring the intrinsic information mining at the sketch and text level.

\vspace{-1mm}
\subsection{Diffusion Models for 3D Shape Generation}
\vspace{-1mm}

Recent methods~\cite{diffusion1,zhou20213d,zeng2022lion_text23d,poole2022dreamfusion_text23d} generate 3D shapes with denoising diffusion probabilistic models. DPM~\cite{diffusion1} introduced the flow-based model for point cloud embedding that conditions the diffusion model on geometry representation. PVD~\cite{zhou20213d} trains a diffusion model on the point-voxel representation for 3D shapes. It performs the variational lower bound optimizing on unconditional shape generation and conditioned shape completion to produce complete 3D shapes. These methods show great promise in extending the diffusion theory to 3D shape generation and achieving good performance. However, their generations are relatively uncontrolled and lack certainty for our generation problem which needs to follow alignment with user instructions more closely.
LION~\cite{zeng2022lion_text23d} proposed a text-based 3D generation model by combining the CLIP-forge to align the text-to-3D embedding. DreamFusion~\cite{poole2022dreamfusion_text23d} is a generative method for text based 3D rendering, which replaces CLIP with a loss derived from the distillation of 2D diffusion models. Point-E~\cite{nichol2022pointe} starts from a language prompt to generate the single view image with fine-tuned GLIDE model and then uses the generated image for image-to-shape diffusion.
Nonetheless, these text-based 3D diffusion models rely on feature pre-alignment from text to image and usually fail when there are huge domain gaps.
Our method focuses on information mining for 3D diffusion by exploring the 3D structure from sketch and text 
to facilitate efficient controlled generation of high quality 3D shapes.


\begin{figure*}[!t] 
		
		\begin{tikzpicture}[inner sep=0pt,outer sep=0pt]
		\centering
		\node[anchor=south west] (A) at (0in,0in)
		{\includegraphics[width=\textwidth,clip=false]{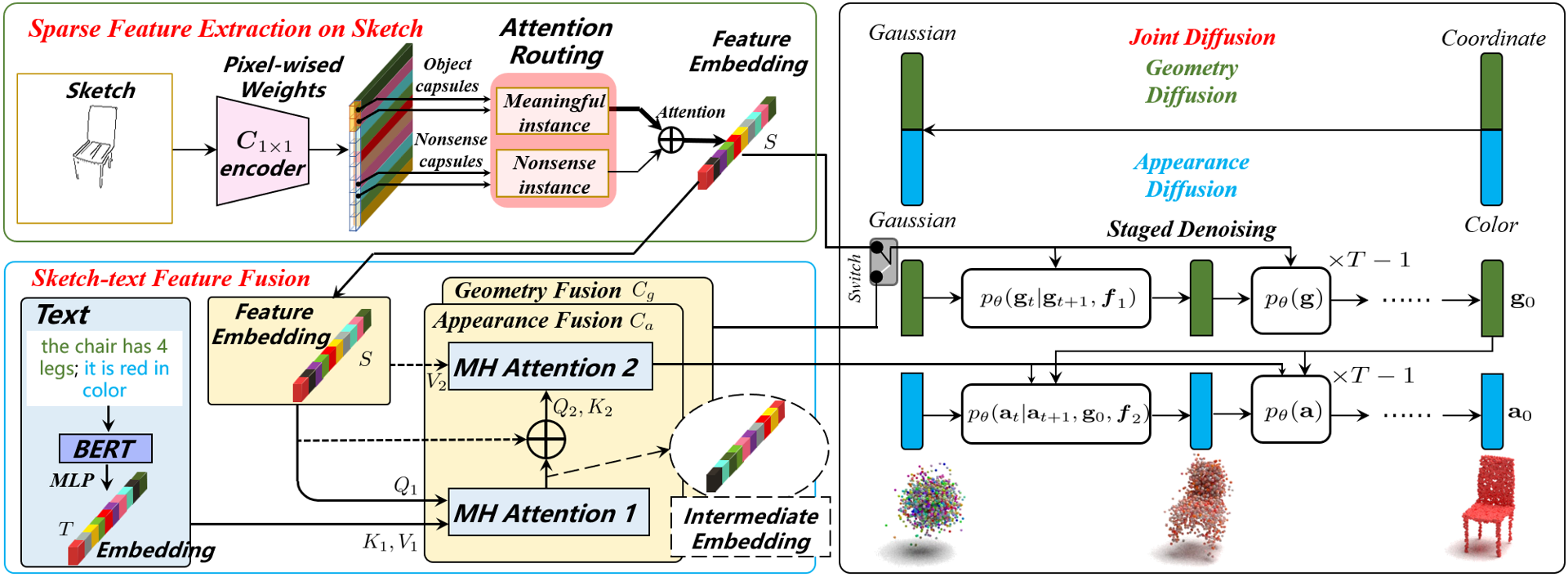}};

		\end{tikzpicture}

		\caption{\label{fig:framework} Overview of STPD framework. The sketch feature is first extracted with two level attention modules for sparse embedding and then flows with multi-head attention fusion to text. With the embedded features, we proposed a staged denoising probabilistic diffusion model for both the geometry and appearance of 3D shapes. }
		\vspace{-3mm}
	\end{figure*}

\vspace{-2mm}
\section{Proposed Approach}
\vspace{-2mm}

Given a sketch $\mathbb{S}$ and text description $\mathbb{T}$ of an object, the goal is to generate its 3D geometry and appearance. To achieve this, the proposed STPD model consists of three modules: sparse feature extraction network for sketch embedding (Section \ref{PAii}), sketch-text feature fusion (Section \ref{PAiii}), and staged diffusion for shape and color generation (Section \ref{PAiv}).
Figure \ref{fig:framework} shows an overview of our model.

		

		





	\begin{figure}[!t] 
		
		\begin{tikzpicture}[inner sep=0pt,outer sep=0pt]
		\centering
		\node[anchor=south west] (A) at (0in,0in)
		{\includegraphics[width=0.48\textwidth,clip=false]{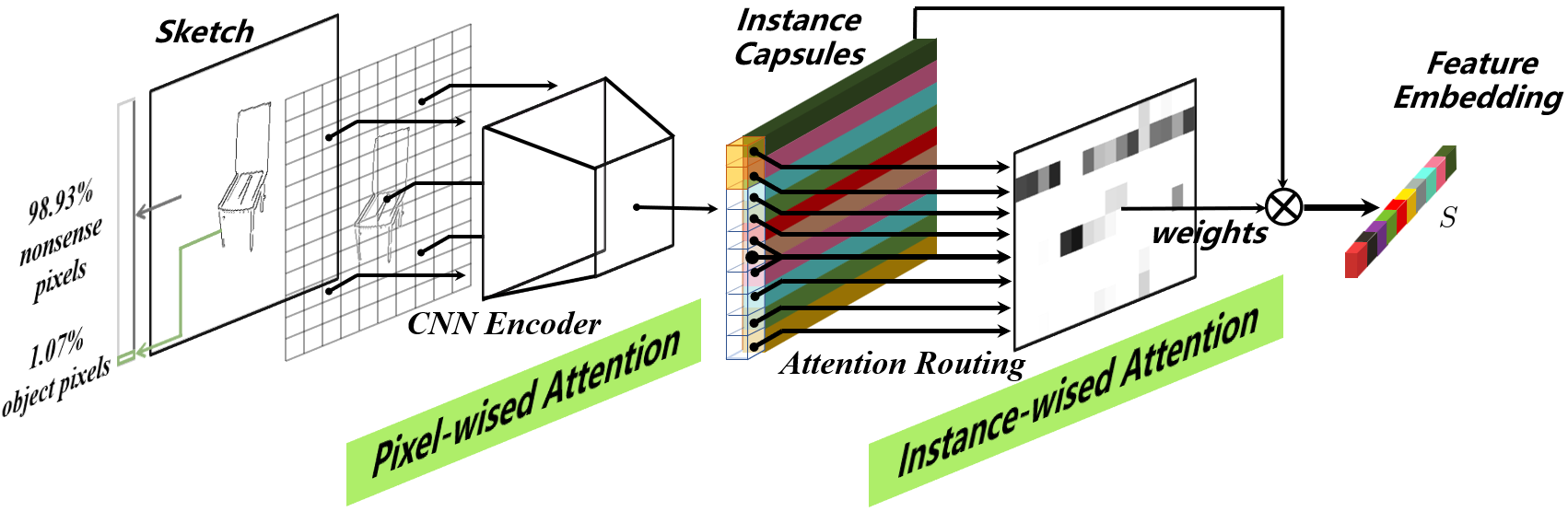}};

		\end{tikzpicture}
		
		\vspace{0mm}
		\caption{\label{fig:sketch} Proposed sketch feature extractor. Pixel-wise attention and instance-wise attention are implemented by a CNN and attention routing, respectively. The two level attentions are designed to extract the sparse information from hand drawn sketches.}
		\vspace{-3mm}
	\end{figure}

\vspace{-1mm}
\subsection{\label{PAii} Sparse Feature Extraction of a Sketch}
\vspace{-2mm}

Feature extraction from a sketch faces three major challenges. Firstly, sketches are sparse and consist of very few informative pixels. Secondly, visual models are generally trained on large amounts of natural images and their attention mechanisms exploit the information in dense pixels to focus on the more relevant parts. This becomes challenging when the image (i.e. sketch) contains copious amounts of futile background pixels and the training dataset is also small. Finally, the relationship between different object parts is ambiguous due to self-occlusions in a single sketch. Below, as shown in Fig.~\ref{fig:sketch}, we present our capsule network using dynamic routing~\cite{sabour2017dynamic_caps} with attention to effectively learning the sparse features from a sketch image.

\noindent \textbf{Primary Capsule Layer:} Our sketch extraction starts from a pixel-wise CNN embedding $S_{N,D_{input},W,H}=\boldsymbol{f}_{cnn}(\mathbb{S})$, where $N$ is the core number of caps, $D$ is the dimension of each caps, and $W, H$ are the width, height of the sketch. Based on these embeddings, we perform a $1\times 1~Conv$ of filter size $D$ to share an affine transformation on each capsule channel and then turn it to a primary capsule $u^{i,0}_{N,D,w,h}$

\vspace{-3mm}
\begin{equation}
    u^{i,0}_{(N,D,w,h)} = {\rm \text{Conv}}(S_{(N,D,w,h)}).
\end{equation}
\vspace{-4mm}

\noindent Here, $i$ denotes the index of capsules in the current layer, $w,h$ is the size of current caps layers and Conv is a simple network comprising a $1\times 1$ convolution layer with stride 1, followed by batch normalization, ReLU, and max pooling. 

\noindent \textbf{Dynamic Routing with Attention:} 
Dynamic routing is applied to strengthen the structural relationship inference between different sketch parts. This facilitates robust feature learning from sketches while keeping the network depth in control and the number of parameters low. We add an attention module to force the capsules to focus on the useful sketch parts and ignore the background.  Each $s^j$ can be acquired by the sum of previous layer capsules,
\vspace{-5mm}
\begin{equation}
    s^{j}_{(N,D,w,h)} = \sum_{i=1,2,\dots} c^{ij}_{w,h} a^{ij}_{(N,D,1,1)} u^{i,j-1}_{(N,D,w,h)},
\end{equation}
\vspace{-5mm}

\noindent where $c^{ij}$ are the dynamic routing parameters computed by softmax layers and $a^{ij}$ is the attention weight obtained by $1\times1$ Conv on lower level primary capsules and hence associated with each instance of the original sketch. 

Similar to the $l$ times dynamic routing, the capsules are squashed~\cite{sabour2017dynamic_caps}, to cluster the informative parts, 
and then 
reshaped to a one dimension vector $S_{(1,D_f)}$ (where $D_f=N\times D$) from the last $s^{j}$. This $S$ forms the sketch embedding used to condition the reverse diffusion process.

\vspace{-1mm}
\subsection{\label{PAiii} Sketch-Text Feature Fusion Embedding}
\vspace{-1mm}

We encode the input language expression as a sequence of textual embeddings with BERT~\cite{devlin2018bert}. Applying a linear projection to the BERT embedding, we get embedding $T$ of the same dimension as the sketch embedding $S$. Both embeddings are fused with two multi-head attention modules to guide the reverse diffusion process. 

The attention-ed embedding is formulated as a set of cascaded multi-attention modules. The first attention takes the sketch feature $S$ as query to extract the relevant {\em intermediate} semantic information $I$ from the text embedding. The second attention performs text-guided self-attention for the sketch feature to produce the conditional embedding that contains geometry/color information based on $I$. Architectures for geometry and color diffusion are the same but have different parameters. 
The first multi-head attention (MH Attention 1 in Fig. \ref{fig:framework}) is defined as:

\vspace{-5mm}
\begin{equation}
       I_{(1,D_f)}  = Atten_1 (Q_1 = S,K_1 = V_1 = T),
\end{equation}
\vspace{-5mm}

\noindent where $Q_1,K_1,V_1$ are the Query, Key, Value and $I$ is an intermediate attention-ed feature passed to the second multi-head attention module (MH Attention 2) defined as:

\vspace{-5mm}
\begin{small}
\begin{equation}
 C_{p(1,D_f)} = Atten_2 ( Q_2 = K_2 = I+S, V_2 = S), 
\end{equation}\end{small}
\vspace{-5mm}

\noindent where  $Q_2,K_2,V_2$ are the Query, Key, Value and $p = g$ for geometric diffusion and $p=a$ for appearance diffusion.  
Specifically, we take the sum of intermediate features $I$ with the sketch feature $S$ as the input of the query and key in the second multi-head attention. Then, the text prompt is introduced as the vector input to support the appearance denoising using the color information from the attention module. Note that, $C_g$ and $C_a$ are from the same cascaded attentions but focus on different descriptions (geometry or appearance) in each diffusion process. Appearance condition also requires information from sketch (shape) because different object parts can have different colors.

\vspace{-1mm}
\subsection{\label{PAiv} Staged Colored Point Cloud Generation}
\vspace{-2mm}

\noindent \textbf{Markov's chain:} 
Given a sample $\mathbf{x}_0\sim q(\mathbf{x}_0)$, noise is systematically added to $\mathbf{x}_0$ in the forward process~\cite{ho2020denoisingdiffusion1} step-by-step following the Markov's chain assumption until the Gaussian distribution is reached:

\vspace{-3mm}
\begin{equation}
\begin{gathered}
q\left(\mathbf{x}_{0: T}\right)=q\left(\mathbf{x}_0\right) \prod_{t=1}^T q\left(\mathbf{x}_t \mid \mathbf{x}_{t-1}\right),\\
q\left(\mathbf{x}_t \mid \mathbf{x}_{t-1}\right) =\mathcal{N}\left(\sqrt{1-\beta_t} \mathbf{x}_{t-1},~ \beta_t \mathbf{I}\right),
\end{gathered}
\end{equation}
\vspace{-3mm}

\noindent where $\mathcal{N}\left( \mu,\sigma^2 \right)$ denotes a Gaussian distribution. Here, $\mathbf{x}_0$ can represent either coordinates or colors from 3D shapes. We train geometry and appearance diffusion separately to generate colored point clouds.
The reverse process, $p_\theta\left(\mathbf{x}_{t} \mid \mathbf{x}_{t+1}\right)$, starts from a standard Gaussian prior and ends with the desired $\mathbf{x}_{0}$:

\vspace{-5mm}
\begin{equation} \label{eq:sample}
    \begin{gathered}
p_\theta\left(\mathbf{x}_{0: T}\right)=p\left(\mathbf{x}_T\right) \prod_{t=1}^T p_\theta\left(\mathbf{x}_{t-1} \mid \mathbf{x}_t\right), \\
p_\theta\left(\mathbf{x}_{t-1} \mid \mathbf{x}_t\right) =\mathcal{N}\left(\mu_\theta\left(\mathbf{x}_t, t\right), \sigma_t^2 \mathbf{I}\right).
\end{gathered}
\end{equation}
\vspace{-3mm}

%
\noindent This cross-entropy minimization is designed such that it forces the reverse process to follow the pattern of noise addition of the forward process, and eventually, recover $\mathbf{x}_0$ from the standard Gaussian noise distribution. Hence, the loss function can be derived as:
\vspace{-3mm}
\begin{equation}\label{eq:loss_o}
    L_t = \mathbb{E}_{\mathbf{x}_0,\epsilon}[C \left\| \tilde{\mu}_t( \mathbf{x}_t,\mathbf{x}_0)-\mu_t (\mathbf{x}_t,t)\right\|^2],
\end{equation}
\vspace{-7mm}

\noindent where $C$ is a constant. 

\vspace{1mm}
\noindent \textbf{Geometry Diffusion:} 
Denoting the colored point cloud as $\mathbf{x}_0\in \mathbb{R}^{N\times 6}= \{\mathbf{g}_0\in \mathbb{R}^{N\times 3},\mathbf{a}_0\in \mathbb{R}^{N\times 3}\}$, consisting $N$ points with geometry $\mathbf{g}_0$ and appearance $\mathbf{a}_0$ coordinates. 
We first perform geometry diffusion to generate the 3D shape. \names is trained by minimizing the cross-entropy loss between two diffusion chains with respect to the geometric conditional feature $C_g$, and learning of model parameters $\theta$:

\vspace{-8mm}
\begin{equation} \label{eq:geometry}
\begin{gathered}
          \min_\theta \mathbb{E}_{g_0\sim q(g_0), g_{1:T}\sim q(g_{1:T})}[\sum_{t=1}^T\log p_\theta(\mathbf{g}_{t-1}|\mathbf{g}_t, C_g)],       
\end{gathered}
\end{equation}
\vspace{-5mm}

\noindent where $C_g$ is derived from sketch-text embedding when the text also contains shape description, or only from sketch when there is no text. We add a switch to adjust the feature map selection for different scenarios.   
The joint posterior $q(g_{1:T})$ can be factorized by the assumption of Markov's chain $\prod_{t=1}^T q(\mathbf{g}_{t-1}|\mathbf{g}_{t},\mathbf{g}_{0})$, where each is analytically tractable by a re-parameterized Gaussian distribution:

\vspace{-4mm}
\begin{equation}
\begin{aligned}
\tilde{\boldsymbol{\mu}}_t &=\frac{\sqrt{\alpha_t}\left(1-\bar{\alpha}_{t-1}\right)}{1-\bar{\alpha}_t} \mathbf{x}_t+\frac{\sqrt{\bar{\alpha}_{t-1}} \beta_t}{1-\bar{\alpha}_t}\mathbf{x}_0, 
\end{aligned} \end{equation}
\vspace{-7mm}

\begin{equation}
q\left(\mathbf{g}_{t-1} \mid \mathbf{g}_t,\mathbf{g}_{0}\right) =\mathcal{N}\left(\mathbf{g}_{t-1};\tilde{\boldsymbol{\mu}}_t, \tilde{\beta}_t \mathbf{I}\right),
\end{equation}

\noindent where $\tilde{\beta}_t$ is the known variance defined in the forward process, $\alpha_t=1-\beta_t$, and $\tilde{\alpha}= \prod_{s=1}^t\alpha_s$. From Eq.~\ref{eq:loss_o} and this property, our diffusion model learns to match each $q\left(\mathbf{g}_{t-1} \mid \mathbf{g}_t,\mathbf{g}_{0}\right)$ and $p_\theta\left(\mathbf{g}_{t-1} \mid \mathbf{g}_t\right)$
by estimating the output noise $\mathbf{\epsilon}_\theta(\mathbf{g}_t,C_g,t)$ to match noise $\mathbf{\epsilon}$:

\vspace{-3mm}
\begin{equation}
\left\|\mathbf{\epsilon} - \mathbf{\epsilon}_\theta(\mathbf{g}_t,C_g,t) \right\|^2, \epsilon\sim \mathcal{N}(0,\mathbf{I}),
\end{equation}
\vspace{-4mm}


The noise prediction network $\epsilon_\theta$ progressively generates the 3D shape using the reverse chain


\vspace{-6mm}
\begin{equation}
    g_{t-1}=\frac{1}{\sqrt{\alpha_t}}\left(\mathbf{g}_t-\frac{1-\alpha_t}{\sqrt{1-\bar{\alpha}_t}} \mathbf{\epsilon}_\theta(g_t,C_g,t)\right)+\sqrt{\beta_t}\mathbf{\epsilon}. \notag
\end{equation}
\vspace{-3mm}

\noindent \textbf{Appearance Diffusion:} Appearance diffusion is performed 
while freezing the geometry parameters. The optimization function, in this case, can be simply modified from Eq.~\ref{eq:geometry} to learn a conditional generative model for appearance:

\vspace{-7mm}
\begin{equation}
     \min_\theta \mathbb{E}_{a_0\sim q(a_0), a_{1:T}\sim q(a_{1:T})}[\sum_{t=1}^T\log p_\theta(\mathbf{a}_{t-1}|\mathbf{a}_t, p_\theta(\mathbf{g}_0), C_a)] \notag
\end{equation}
\vspace{-5mm}

\noindent The training loss for appearance diffusion is minimized and conditioned on appearance features $C_a$ and $p_\theta(\mathbf{g}_0)$:

\vspace{-3mm}
\begin{equation}
    \left\|\mathbf{\epsilon} - \mathbf{\epsilon}_\theta(\mathbf{a}_t,p_\theta(\mathbf{g}_0),C_a,t) \right\|^2, \epsilon\sim \mathcal{N}(0,\mathbf{I}).
\end{equation}
\vspace{-5mm}

\noindent \textbf{Colored Point Cloud Generation:} 
We can now formally define our joint generative model for colored point cloud generation as 

\vspace{-5mm}
\begin{small}
\begin{equation}
    p_{\theta_2,\theta_1,\phi}(\mathbf{a}_0,\mathbf{g}_0,C_g,C_a)= p_{\theta_2}(\mathbf{a}_0|\mathbf{g}_0,C_a)p_{\theta_1}(\mathbf{g}_0|C_g)p_\phi(C_g,C_a), \notag
\end{equation}
\end{small}
\vspace{-5mm}

\noindent where $\{\mathbf{a}_0,\mathbf{g}_0\}$ denote the colored point cloud $\mathbf{x}_0$. $p_\phi(C_g,C_a)$ refers to the sketch and textual embedding, and $p_{\theta_2}(\mathbf{a}_0)$, $p_{\phi}(\mathbf{g}_0)$ are the appearance and geometry denoising processes respectively. Note that the appearance diffusion is conditioned on the geometry diffusion as well.
We also tried extended diffusion to denoise from a 6D Gaussian rather than treating the shape and appearance separately. However, integrating the unrelated features causes undesirable interference in the diffusion process and consequently, \names does not learn to produce point clouds with fine details.



\vspace{-2mm}
\section{Experiments}
\vspace{-1mm}

\begin{figure*}[!t] 
	
	\begin{tikzpicture}[inner sep=0pt,outer sep=0pt]
	\centering
	\node[anchor=south west] (A) at (0in,0in)
	{\includegraphics[width=\textwidth,clip=false]{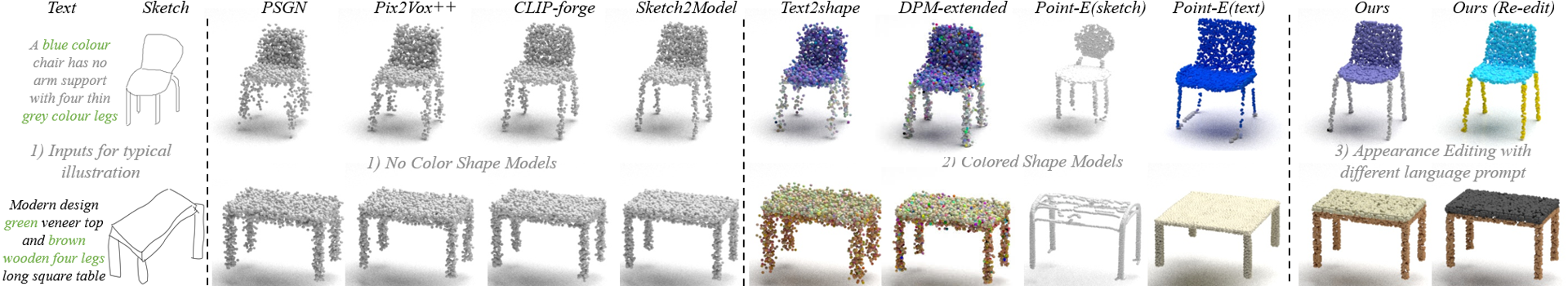}};		
	\end{tikzpicture}	
	\caption{\label{fig:vis} Visual comparison of the generated objects.  1) Inputs include text and free-hand sketch. 2) Shape methods with no color. 3) Colored shape generation methods. 4) Our method generates both geometry and appearance using sketch and text. The appearance can be re-edited by changing the language prompt while preserving geometry.}
	\vspace{-5mm}
\end{figure*}

\subsection{Experimental Setup}
\vspace{-1mm}
\noindent \textbf{Datasets:} We use the $chair,~table,~aeroplane,~car$ categories of ShapeNet dataset \cite{chang2015shapenet}, randomly split it into training-test sets at 80-20\% ratio and render a single-image from each 3D object. We apply Canny edge detection~\cite{Canny} on the rendered ShapeNet images to acquire their single-channel edge images. We manually remove the background so that they resemble clean hand drawn sketches. For colored point clouds, we use the $chair,~ table$ categories as only they have text descriptions~\cite{chen2018text2shape}. To generate ground-truth point clouds, each 3D shape is randomly sampled at 2048 points and then translated to zero mean and normalized in the range $[-1,1]^3$ so that the Gaussian distribution assumption is satisfied during the forward pass.

\begin{table}[htbp]
  \footnotesize 
	\centering
	\addtolength{\tabcolsep}{-4.65pt}
	
	\begin{tabular}{c c  l| c c | c c }
		\toprule
		\multirow{2}{*}{Categories} & \multirow{2}{*}{Input} & \multirow{2}{*}{Methods}  & \multicolumn{2}{c}{MMD ($\downarrow$)}	&	\multicolumn{2}{c}{COV ($\%,\uparrow$)}\\ 
		    &  & & CD  & EMD & CD  & EMD    \\
		\midrule
        \multirowcell{6}[0pt][c]{Reconstruction\\Shape (N$\times$3)} & \multirowcell{4}[0pt][c]{Sketch} & PSGN~\cite{PSGN}  & 8.174  & 7.253 &64.13 &69.76 \\
            && Pix2Mesh~\cite{Pix2mesh} & 9.771&7.347&65.61&67.86 \\
           & & TMNet~\cite{TMnet} &9.18&8.11&63.55&65.24 \\
           & &Sketch2Mesh~\cite{Sketch2mesh} & 3.018 & 2.786 & 80.24 & 78.23\\ \cmidrule{2-7}
        & \multirowcell{2}[0pt][c]{Sketches} & Pix2Vox++~\cite{pix2vox}   &  5.976& 5.760& 71.29&69.55 \\
           & & EvolT ~\cite{EvolT}    & 5.718& 5.307& 73.09&74.84 \\ \hline  
         \multirowcell{7}[0pt][c]{Generative\\Shape (N$\times$3)} & \multirowcell{2}[0pt][c]{Text} &   Text2Shape ~\cite{text2model2}	&	9.873	&	6.348	& 33.76 &31.28\\
         &  & CLIP-forge ~\cite{text2model1}	&	4.367	&   3.818	& 35.26 & 33.17\\ \cmidrule{2-7}
            & \multirowcell{3}[0pt][c]{Sketch} &	MixNF~\cite{image2model1}	    &	7.689	&	6.224	& 69.89 &67.24\\
	      &  &	DPF-Net ~\cite{image2model2}	&	7.531	&	6.176	& 72.34 & 67.68\\
		  &  &   Sketch2Model~\cite{sketch2model}	        &	3.176	&	2.967   & 78.69 & 77.82\\  \cmidrule{2-7}
		  & \multirowcell{1}[0pt][c]{$\times$} &   DPM ~\cite{diffusion1}	    &	3.764	&	1.986	& 46.87 & 44.59\\\cmidrule{2-7}
		  & \multirowcell{1}[0pt][c]{Sketch,text} &   STPD (ours)		                    &	\textbf{1.635}	&	\textbf{1.043} & \textbf{93.53}	 & \textbf{91.37} \\ \hline 
      \multirowcell{5}[0pt][c]{Generative\\Shape\&Color \\(N$\times$6)} & \multirowcell{2}[0pt][c]{Text}
	        &   Text2Shape & 10.34 & 7.476 &  34.59& 33.26 \\
       &  &   Point-E (text)~\cite{nichol2022pointe} & 2.523 & 2.265& 51.96 & 49.23\\ \cmidrule{2-7}
        & \multirowcell{1}[0pt][c]{$\times$}  &DPM-extended                & 4.765 &3.256 &	45.76& 39.68\\\cmidrule{2-7}
        & \multirowcell{1}[0pt][c]{sketch}  &   Point-E (sketch) & 3.092 & 3.117& 61.12 & 62.09\\\cmidrule{2-7}
         & \multirowcell{1}[0pt][c]{Sketch,text}  &   STPD (ours)		           & \textbf{1.346} & \textbf{1.017} & \textbf{95.53}& \textbf{94.97}\\
		\bottomrule
	\end{tabular}
        \caption{\label{tab:comparison1} Comparison of 3D shape generation for the $chair,~table$ categories of ShapeNet \cite{chang2015shapenet}. CD is multiplied by $10^3$ and EMD by $10^1$. Our method generates the best shape, with or without color. }
 \vspace{-5mm}
\end{table}

\noindent \textbf{Evaluation Metrics:} Previous works \cite{diffusion1} used Chamfer Distance (CD) and Earth Mover's Distance (EMD) to evaluate the reconstruction quality of point clouds. Based on these, existing works report Coverage (COV) and Minimum matching distance (MMD). Although these metrics have their limitations, they still serve as a good reference to measure and compare the performance of generative methods. 
Specifically, COV evaluates the certainty of conditioned generation whereas MMD calculates the average distance of the point clouds in the reference set and their closest neighbors in the generated set. Another metric used in numerous works to evaluate the generation quality and diversity is 1-nearest neighbor accuracy (1-NNA) which directly quantifies distribution similarity between two sets. 
However, since our aim is to generate 3D shapes faithful to the provided sketch, 
we stick to the first four metrics. Where other methods do not provide color, we compare shapes only.



\noindent \textbf{Baselines:} We compare with classical as well as diffusion based methods. In the former category, we select the major sketch~\cite{sketch2model}, image~\cite{image2model1}~\cite{image2model2}, and text~\cite{text2model1}~\cite{text2model2} based 3D object generation methods. We also compare to reconstruction based methods including single-view~\cite{PSGN,Pix2mesh,TMnet,Sketch2mesh} and multiviews~\cite{xie2020pix2vox++,EvolT}.
In the latter category, we select 3D diffusion (DPM)~\cite{diffusion1}, Point-E~\cite{nichol2022pointe}, and Point Voxel Diffusion (PVD) \cite{zhou20213d}. Point-E (sketch) results are obtained after fine-tuning the Point-E image-to-shape diffusion model on our sketch-shape data for a fair comparison.
We take ten generations for all methods and use the average error from ground truth for evaluation. 

\vspace{-2mm}
\subsection{Comparisons with State-of-the-art Methods}
\vspace{-2mm}
Table~\ref{tab:comparison1} compares the point cloud generation quality of our method with existing state-of-the-art.  For a fair comparison, we re-train all methods with our sketch-shape data of the $chair,~table$ categories, or their text descriptions when the method takes text only as input. Multiview methods are given sketches from 3 viewpoints. Only our method performs (single view) sketch and text guided 3D object generation. Since DPM \cite{diffusion1} is unconditional, we generate many shapes and pick the best matches. For shape only (upper half of Table \ref{tab:comparison1}), the proposed STPD outperforms all methods, achieving 45.8\% lower CD-MMD error from the nearest reconstruction method Sketch2Mesh \cite{Sketch2mesh} and 48.5\% lower CD-MMD error from the nearest generative method Sketch2Model \cite{sketch2model}. The multiview methods do not perform well since they are unable to establish correspondences between the sketches. Text guided generative methods have more diversity but struggle with alignment to user specifications given the inherent shape ambiguity in the text. Sketch guided methods perform better in general.

Table \ref{tab:comparison1} also compares colored point cloud (shape and color) generation. For this, we extend DPM to a 6D Gaussian and also compare with Text2Shape \cite{chen2018text2shape} and Point-E \cite{nichol2022pointe}. The proposed STPD again outperforms all methods, achieving 46.7\% CD-MMD error from the nearest competitor Point-E \cite{nichol2022pointe}. In Fig.~\ref{fig:vis}, we compare our results visually to the existing methods using hand-drawn sketches from SketchX~\cite{li2014fine}. Our method also offers to re-edit the color (appearance) once the geometry is finalized (last column). 
Table~\ref{tab:comparison_3c} reports comparative results on the $chair,~table,~aeroplane,~car$ categories of ShapeNet. We choose these categories because PVD \cite{zhou20213d} provides comparisons on the $chair,~aeroplane,~car$ categories and the $table$ category is used in our previous experiment. From Table \ref{tab:comparison_3c}, we can see that STPD achieves the best performance on both evaluation metrics on all four object categories. SoftFlow \cite{kim2020softflow}, PointFlow~\cite{yang2019pointflow}, DPF-Net~\cite{image2model2}, DPM~\cite{diffusion1}, Point-E~\cite{nichol2022pointe}, PVD~\cite{zhou20213d} achieve an average Chamfer Distance of 2.464, 2.429, 2.168, 2.509, 2.182, 2.539 and an average EMD of  3.105, 3.191, 2.851, 2.553, 2.490, 2.661 respectively. Our STPD achieves 1.172, 1.261 average CD, EMD errors which are 46.3\%, 49.3\% lower than the nearest competitor Point-E \cite{nichol2022pointe}.

\noindent \textbf{Human Evaluations:} We also conducted human evaluations on 100 objects generated by our method and other methods. Table~\ref{Human_rate} shows the results where 30 people rate the generations on a scale of 1 to 5 (higher is better). \names outperforms the baseline methods on all individual object categories and achieves an average rating of 4.4, compared to average rating of 3.3 by the nearest competitor Point-E.

\begin{table}[t]
        \footnotesize
	\centering
	\addtolength{\tabcolsep}{-3.8pt}
	
	\begin{tabular}{l | c c | c c | c c |c c}
		\toprule
  \multirow{2}{*}{Methods}        &    \multicolumn{2}{c|}{chair}    &    \multicolumn{2}{c|}{table} &    \multicolumn{2}{c|}{aeroplane} &    \multicolumn{2}{c}{car} \\ \cmidrule{2-9}
                    &   CD              &   EMD    &   CD              &   EMD   &   CD              &   EMD   &   CD              &   EMD        \\
                    
		\midrule
		  SoftFlow~\cite{kim2020softflow}             & 2.786       &	3.295    & 4.815 & 5.139& 0.404        &	1.198   & 1.850        &	2.789\\
		  PointFlow~\cite{yang2019pointflow}      	& 2.707       &	3.649	 	& 4.804 & 5.082 & 0.403        &	1.180& 1.803        &	2.851\\
		  DPF-Net~\cite{image2model2}       	& 2.763        &	3.320  	&3.986 &4.661		& 0.528        &	1.105& 1.396        &	2.318\\
            DPM~\cite{diffusion1} & 3.201& 2.718 & 4.601 & 4.298 & 0.432  & 1.027&  1.804&2.167\\
            Point-E~\cite{nichol2022pointe}& 2.901 & 2.667  & 3.809 & 4.290  &0.412& 1.005& 1.609& 1.997\\
		  PVD ~\cite{zhou20213d}       	& 3.211        &	2.939		&4.731 &4.527	& 0.442        &	1.030& 1.774        &	2.146\\	\hline
		  STPD (ours)     		& \textbf{1.209}        &	\textbf{1.117}  	&\textbf{2.117} &\textbf{1.897} & \textbf{0.311}        &	\textbf{0.897}& \textbf{1.049}   &	\textbf{1.136}\\

		\bottomrule
	\end{tabular}
        \caption{\label{tab:comparison_3c} Comparison with baselines on 4 categories of the ShapeNet dataset \cite{chang2015shapenet}. CD is multiplied by $10^3$ and EMD by $10^1$.}
        \vspace{-1mm}
\end{table}

\begin{table}[t]
    \footnotesize
	\centering
	\addtolength{\tabcolsep}{-5pt}
	\vspace{-0mm}
	\begin{tabular}{l | c c c c c c | c}
		\toprule [0.1pt]
		Human rate ($\uparrow$)         &  	   SoftFlow           &   PointFlow & DPF-Net & DPM & Point-E & PVD & STPD\\[-0.4ex]
		\midrule 
		chair          &  	   1.8 & 1.7 & 1.9 & 2.1& 3.9 & 3.5 & \textbf{4.8}  \\[-0.4ex]
        table          &  	   2.1 & 1.9 & 3.5 & 3.0 & 3.4 & 2.5 & \textbf{4.6}\\[-0.4ex]
        aeroplane          &   3.4 & 3.3 & 2.2 & 3.1 & 3.7 & 3.0 & \textbf{4.0}\\[-0.4ex]
        car          &  	   2.5 & 2.4& 3.4 & 3.5 & 3.2 & 2.9 & \textbf{4.3}\\[-0.4ex]
		\bottomrule
		
	\end{tabular}
 \caption{\label{Human_rate} {Human evaluations: 100 generated objects rated on a scale of 1 to 5 by 30 people.}}
 \vspace{-1mm}
\end{table}




\begin{table}[t]
        \footnotesize
	\centering
	\addtolength{\tabcolsep}{-2pt}
 
	\begin{tabular}{l l | c c }
		\toprule
		\multicolumn{2}{c|}{  \underline{Input Modality/Model for Conditioning }}  & CD	& EMD	\\
		Sketch             & Text           &  	     ($\times10^3$)           &   ($\times10^1$) \\
		\midrule
		$\times$ 	&	    BERT                        & 	28.325	    &	20.457	\\
        C$_{1\times 1}$	                &	      $\times$           & 10.965        &	8.658		\\	
		C$_{1\times 1}$                 &  BERT                 &   9.651      &       8.459    \\ \hline
        ResNet50  &             BERT                & 6.409      &  5.480 \\
        ResNet50 + Atten.Routing &  BERT          & 1.935     &     1.587 \\
        C$_{1\times 1}$ + Atten.Routing	& 	   $\times$          & 	1.689	    &	1.348		\\\hline
        C$_{1\times 1}$ + Atten.Routing (concatenate) & BERT            & 	3.748	        &	4.127		\\
		C$_{1\times 1}$ + Atten.Routing (our fusion) & BERT	           & 	\textbf{1.390}	        &	\textbf{1.101}		\\
		\bottomrule
	\end{tabular}
 \caption{\label{tab:comparison2} Ablation study on visual \& textual modules. C$_{1\times 1}$ is a simple visual encoder comprising ${1\times 1}$ convolution followed by batch normalization and ReLU.}
 \vspace{-5mm}
\end{table}

		

\subsection{Ablation Study}
\vspace{-1mm}
We perform ablation study on the sketch and text modules to verify their contributions. 
Table~\ref{tab:comparison2} shows our results using CD and EMD metrics. 
Text conditioning, using BERT embeddings, alone achieves the worst performance.  C$_{1\times 1}$ is a simple visual encoder comprising $1\times 1$ convolution followed by batch normalization and ReLU. Sketch conditioning, using simple C$_{1\times 1}$ embeddings, reduces the errors significantly since a sketch can describe shaping much better than text. Using sketch and text conditioning with C$_{1\times 1}$ with BERT embeddings, the errors do not reduce much. However, when a more sophisticated sketch encoding is performed with ResNet50 \cite{Resnet} and combined with BERT, the CD and EMD reduce notably. To account for the sparsity of sketches, when we add capsule network with attention routing on top of ResNet50, we see a dramatic improvement in both metrics. Our simple visual encoder C$_{1\times 1}$ with attention routing further improves the performance even without using text conditioning. As seen in the last row of the table, the best performance is achieved when we use sketch conditioning with C$_{1\times 1}$ + attention routing embeddings combined with text embeddings from BERT~\cite{devlin2018bert} using the proposed fusion method. The second last row of the table shows that a simple concatenation of the two embeddings does not work as well as our proposed fusion mechanism.

\begin{table}[t]
        \footnotesize
	\centering
	\addtolength{\tabcolsep}{0pt}
	\vspace{-0mm}
	\begin{tabular}{l | c c c}
		\toprule
		Model          &  	   ModelNet10           &   ModelNet40 & {ScanObjectNN}\\
		\midrule
		PointFlow~\cite{yang2019pointflow}            & 93.7        &	86.8	& 78.2	\\
		PDGN~\cite{hui2020progressive_PDGN}	        & 	 	94.2	    &	87.3	&78.1	\\
		3D-GAN~\cite{3DGAN}            &   91.0      &       83.3  & 76.7 \\
		AtlasNet~\cite{AtlasNet}            &   91.9      &       86.6   & 76.9\\
		ShapeGF~\cite{ShapeGF}            &   90.2     &       84.6   & 75.8\\
		DPM~\cite{diffusion1}             & 94.2          &   87.6      & 79.7  \\ \cmidrule{1-4}
		STPD            &   \textbf{94.8}     &       \textbf{88.1}   & {\bf 84.3} \\
		\bottomrule
		
	\end{tabular}
 \caption{\label{tab:comparison4} Object classification accuracy (\%).}
 \vspace{-1mm}
\end{table}

\vspace{-1mm}
\subsection{Representation Learning Evaluation}
\vspace{-2mm}
We evaluate the representation learned by our sketch embedding used to condition the diffusion model. We follow the work of MVCNN~\cite{su2015multi_mvcnn}, ScanObjectNN~\cite{uy-scanobjectnn-iccv19} and convert rendered images from different viewpoints of the ModelNet dataset~\cite{modelnet} to sketches and conduct object classification experiments. We train our STPD on the generated sketches and then take the conditioned embedded features to learn a linear SVM classifier for classification based on a single sketch embedding. The essence of the conditional diffusion model is that the conditioning should guide the diffusion to retrieve the 3D shape from Gaussian noise. This means that our conditional network should be able to represent specific object categories to aid classification. The results in Table \ref{tab:comparison4} demonstrate that our architecture can extract discriminative features, that are also useful for accurate classification.


	

 	\begin{figure}[tpb] 
		
		\begin{tikzpicture}[inner sep=0pt,outer sep=0pt]
		\centering
        \vspace{-5mm}
		\node[anchor=south west] (A) at (0in,0in)
		{\includegraphics[width=0.49\textwidth,clip=false]{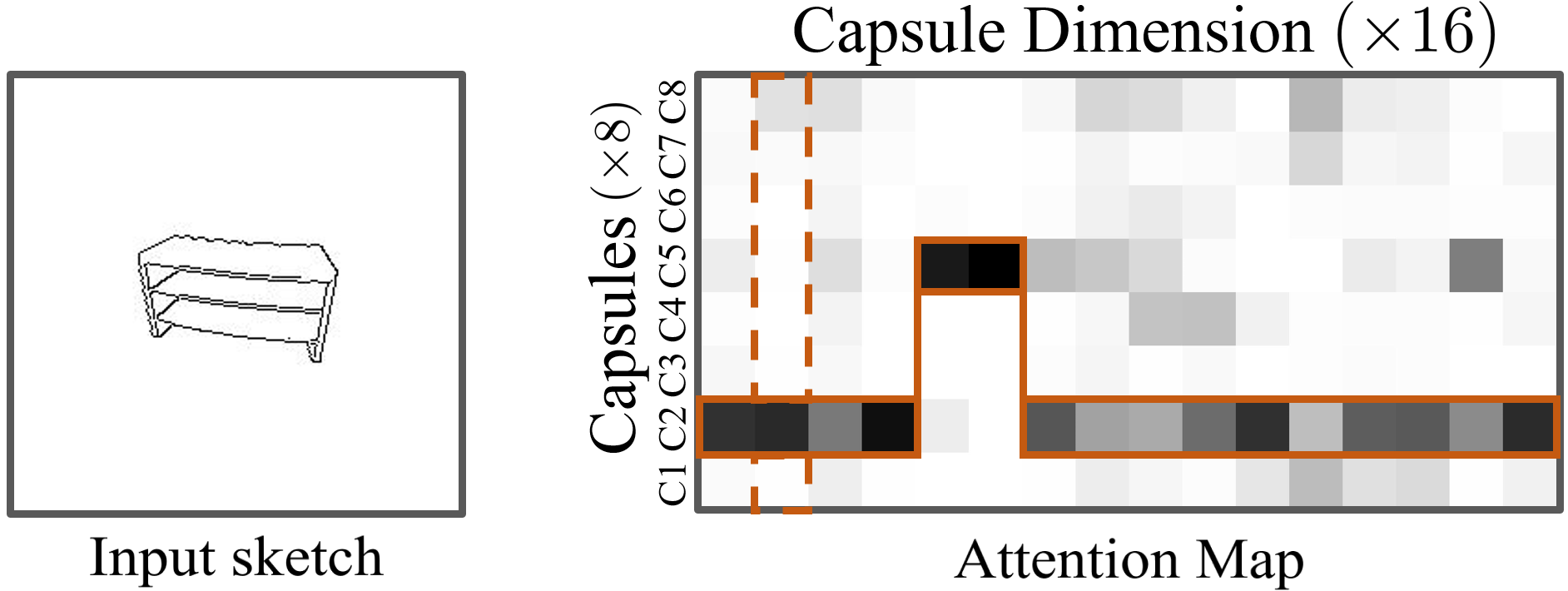}};

		\end{tikzpicture}
		
		\vspace{-1mm}
		\caption{\label{fig:attentionss} Visualization of attention score maps for a sketch with respect to capsules. The 8$\times$ 16 size denotes 8 capsules of 16-dimension each. Capsule attention maps are learned from attention-routing. Dark parts denote high attention. 
        }
		\vspace{-4mm}
	\end{figure}

 \begin{figure*}[h!] 
		
		\begin{tikzpicture}[inner sep=0pt,outer sep=0pt]
		\centering
		\node[anchor=south west] (A) at (0in,0in)
		{\includegraphics[width=0.96\textwidth,clip=false]{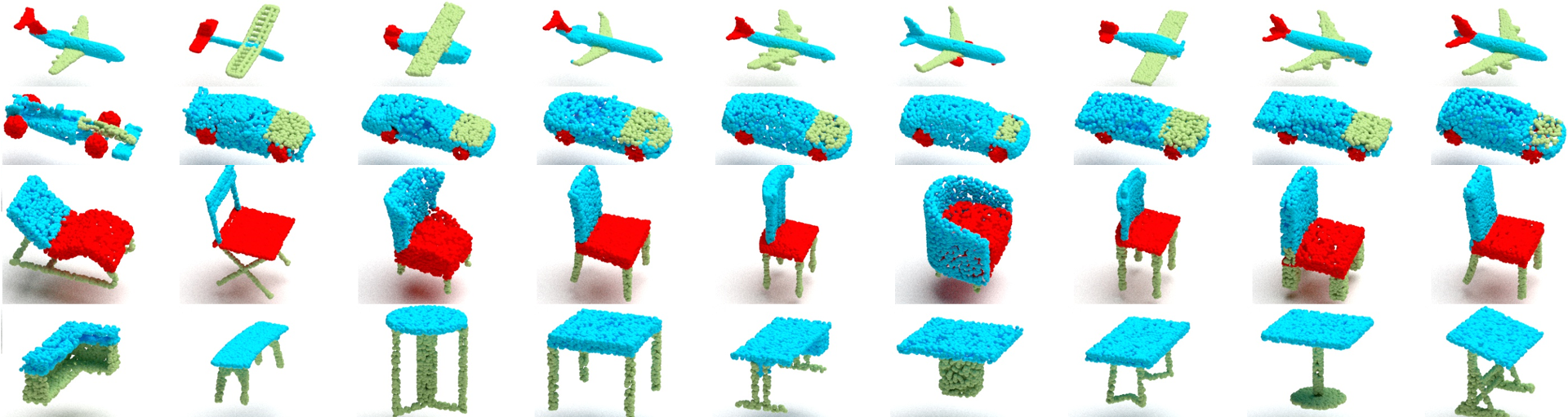}};

		\end{tikzpicture}
		
		\vspace{-0mm}
		\caption{\label{fig:segs}  Visualization of part segmentation results for 4 categories of ShapeNet-Parts.} 
		\vspace{-4mm}
	\end{figure*}

\vspace{-1mm}	
\subsection{Effectiveness of Capsule Network}
\vspace{-2mm}

Fig.~\ref{fig:attentionss} visualizes the generated attention scores from a typical sketch. More examples are given in the supplementary material along with quantitative results using Instance Scores for Sketch (ISS i.e. without attention routing) and Instance Score on Capsule (ISC i.e. with attention routing) metrics. ISS is the percentage of useful pixels divided by the total pixels of the sketch. For ISC, we compute the average of the highest attention score for each of the 16 capsule dimensions (column wise) and convert it to a percentage. 

Before the attention capsule, the sketches have ISS of 1.07\% on average. After our capsule-attention module, ISC becomes 58.68\% on average for better diffusion. {This demonstrates the effectiveness of the Capsule Attention Network which attends to sparse data in sketches during diffusion training. 
}  Each capsule ($1\times16$) denotes a learned feature, e.g. the table, from a sketch. It can be observed that given the distribution of the useful pixel in the sketch, only a few attention score values are dark i.e, attentioned. Caps number 2 and a few dimensions of Caps number 5 are given high weight to contribute to the final visual feature, while the other Caps that represent the white background of the sketch are ignored.
The attention-based capsule network enables our method to effectively locate the referred objects in the sparse sketch input. 

\vspace{-1mm}
\section{Extending STPD to Part Segmentation Task}
\vspace{-0mm}
To further show the feature learning ability of our STPD model, we use it for the task of part segmentation in point clouds. For this, we use the ShapeNet-Parts \cite{chang2015shapenet} dataset, focusing on the four object categories of Table \ref{tab:comparison_3c}. 

The part segmentation task can be extended from our colored point cloud generation model by assigning different colors to each part. We consider only two parts for the $table$ and three parts for remaining three categories e.g. for $aeroplane$, the ($body, ~wing, ~tail$) are assigned  ($blue, ~green, ~red$) colors respectively.

In our \names, each coordinate is generated by our sketch-text guided geometric diffusion model. However, in the part segmentation task, the shape $\mathbf{g}_0$ is given as input to the model.  
%
Here, we directly apply the text feature as the condition $C_a$ of appearance diffusion. For instance, we provide a generic text sentence ``{\em An blue aeroplane with green wings and red tail}'' for all aeroplanes.

With the given input geometry and appearance description, we can formally define our joint generative model for point cloud segmentation as 

\vspace{-5mm}
\begin{small}
\begin{equation}
    p_{\theta,\phi}(\mathbf{a}_0,\mathbf{g}_0,C_a)= p_{\theta}(\mathbf{a}_0|\mathbf{g}_0,C_a)p_{input}(\mathbf{g}_0)p_\phi(C_a),
\end{equation}
\end{small}
\vspace{-3mm}

\noindent where the combination $\{\mathbf{a}_0,\mathbf{g}_0\}$ denotes the desired segmentation of the 3D point cloud. Note that, $\mathbf{a}_0$ is generated as colors which are then used as segmentation labels. We apply K-means clustering for final refinement of the colors.

\vspace{2mm}
\noindent \textbf{Results for 3D Shape Segmentation:} Fig.~\ref{fig:segs} shows some visualizations of our results and Table~\ref{tab:comparisonseg} compares our method to some existing baselines. The proposed STPD achieves the highest mIoU compared to existing methods. 
\begin{table}[t]
        \footnotesize
	\centering
	\addtolength{\tabcolsep}{-3pt}
	
	\begin{tabular}{l | c c | c c c c}
		\toprule
		Model          &  	   mIoU$_{c}$            &   mIoU$_{I}$ & chair & table& aeroplane & car\\
		\midrule
		PointNet~\cite{point_net}          & 82.1      &	83.5	 & 89.6   &	80.6 &83.4  & 74.9\\
		PointNet++~\cite{qi2017pointnet++}        & 	 	83.3	    &	84.6	 & 90.8 & 82.6	& 82.4 & 77.3\\
		DGCNN~\cite{wang2019dynamic}            &   83.8      &       84.9   & 90.6 & 82.6 &    84.0 & 77.8\\
		Point-BERT~\cite{yu2022pointbert}            &   84.7     &       85.6    & 91.0 & 81.5 & 84.3 & 79.8\\
		PointMLP~\cite{ma2022rethinking_pointmlp}            &   84.6     &       84.8   & 90.3 & 84.3& 83.5 & 80.5 \\
		KPConv~\cite{thomas2019kpconv}            & 85.1          &   85.9     &91.1& 83.6  & 84.6 & 81.1 \\ 
        P2P~\cite{wang2022p2p}        & 85.0 & 85.9  & 91.6 &          83.7& 84.3 & 80.4 \\\cmidrule{1-7}
		STPD           &   \textbf{85.9}     &       \textbf{86.7}   & \textbf{92.4}&  \textbf{84.1}& \textbf{85.6} & \textbf{81.3}\\
		\bottomrule
		
	\end{tabular}
 \caption{\label{tab:comparisonseg} Comparative results of part segmentation on four categories of ShapeNet-Parts dataset.}
 \vspace{-6mm}
\end{table}

\section{Conclusion}

We proposed a novel sketch and text guided probabilistic diffusion model for 3D object generation. Our model takes a hand drawn sketch and text as inputs to condition the reverse diffusion process. With the strong generative ability and effective guidance by the sketch-text fused features, our model is able to achieve high alignment with user specifications. Our visual encoder is designed to focus on the informative sparse parts of sketches. 
The sketch-text fusion part gathers informative object context from visual and textual descriptions to improve the shape and appearance of the generated 3D objects.  
Experimental results demonstrate that our model achieves state-of-art performance in 3D object generation and shows promising applications for 3D object classification and part segmentation.

\section{Acknowledgement}

Professor Ajmal Mian is the recipient of an Australian Research Council Future Fellowship Award (project number FT210100268) funded by the Australian Government. 

\newpage
\renewcommand\thesection{\Alph{section}}
\renewcommand\thefigure{\Alph{figure}}
\renewcommand\thetable{\Alph{table}}
\noindent\textbf{Supplementary Material}
Here, we provide more details of our proposed method, including color point cloud generation from freehand sketch+text and from only freehand sketch. Section 2 presents typical attention maps for sketches and compares the instance scores before and after attention. In Section 3, we also include a more detailed comparison between DPM \cite{diffusion1}, PVD~\cite{zhou20213d}, point-e~\cite{nichol2022pointe}, and STPD.

We also discuss some of the interesting features of our proposed method at the end.

\section{Color Point Cloud Generation}
\begin{figure*}[!htpb] 
	
	\begin{tikzpicture}[inner sep=0pt,outer sep=0pt]
		\centering
		\node[anchor=south west] (A) at (0in,0in)
		{\includegraphics[width=0.98\textwidth,clip=false]{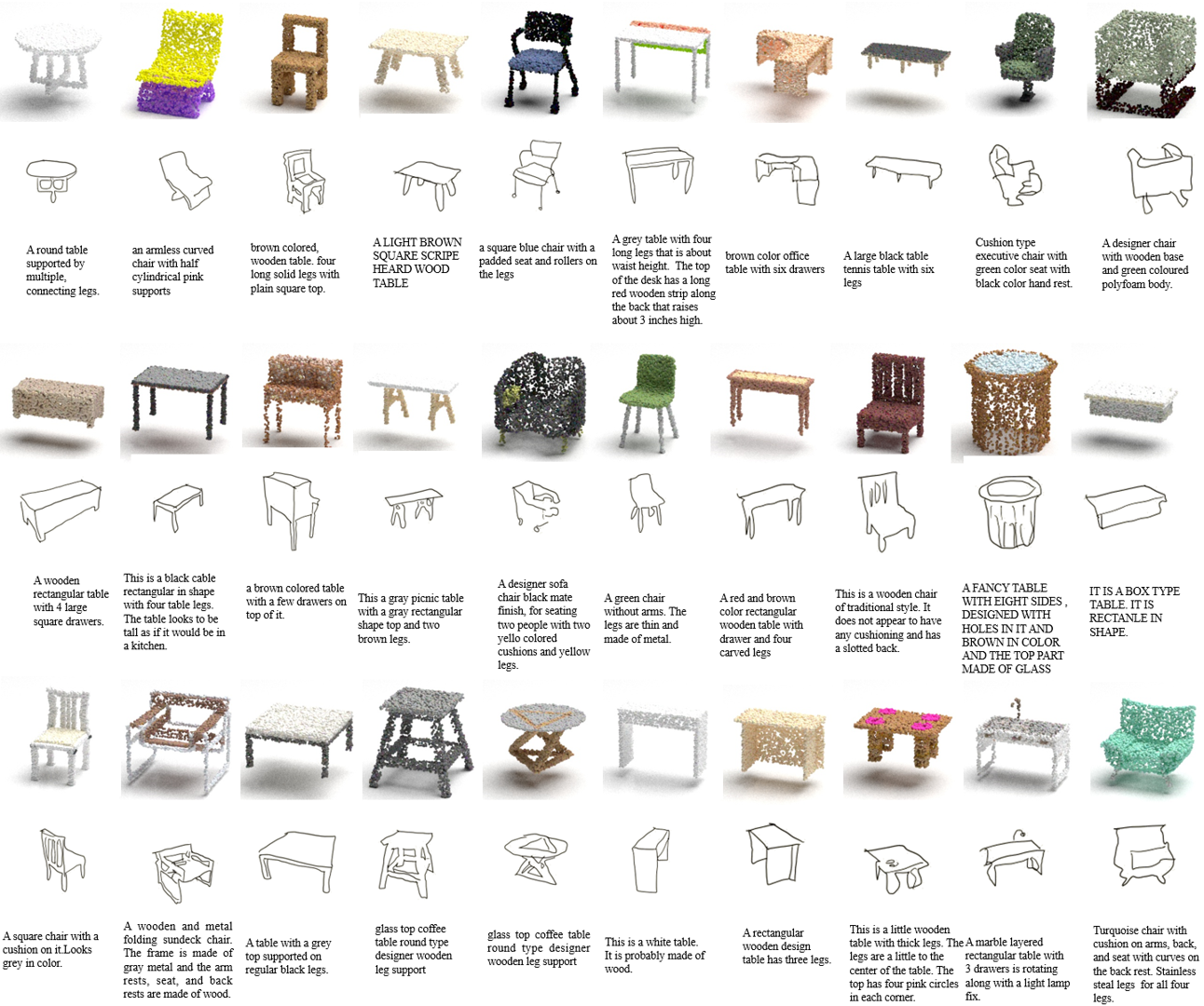}};

	\end{tikzpicture}

	\caption{\label{fig:color} Visualization of the generated objects from free-hand sketch and text. Our model generates better geometry and has the additional functionality of generating colored point clouds. Being conditional, our method produces colored shapes that are more faithful to the input sketch-text descriptions.}
	\vspace{-4mm}
\end{figure*}

Returning to the color point cloud generation ability of our proposed model, in this section, we present additional visualizations for 3D generation of the $chair$ and $table$ categories from freehand sketch and text in point cloud generation (shown in Figure \ref{fig:color}). Specifically, only $chair$ and $table$ categories have color descriptions to support the sketch-text based coloring. We ensure that the edge detection sketches match hand drawn ones. Our conditional method produces colored shapes that are more faithful to the input sketch-text descriptions. Notice how our method pays more attention to the shape details apparent in the sketch (e.g., see the table legs). Note that for better visualization, we up-sample each point cloud to 4096 points.

\section{Typical Attention Maps Related to Sketches}
We also present additional attention maps generated by our attention-capsule network. In this section, we introduce the Instance Score as the evaluation metric. For the instance score on sketch (ISS), we count the number of useful pixels and express it as a percentage of all sketch pixels. For the instance score on capsule (ISC), we compute the highest attention score of each dimension to all capsules and use their average as the evaluation metric.

\begin{figure*}[!htpb] 
	
	\begin{tikzpicture}[inner sep=0pt,outer sep=0pt]
		\centering
		\node[anchor=south west] (A) at (0in,0in)
		{\includegraphics[width=0.98\textwidth,clip=false]{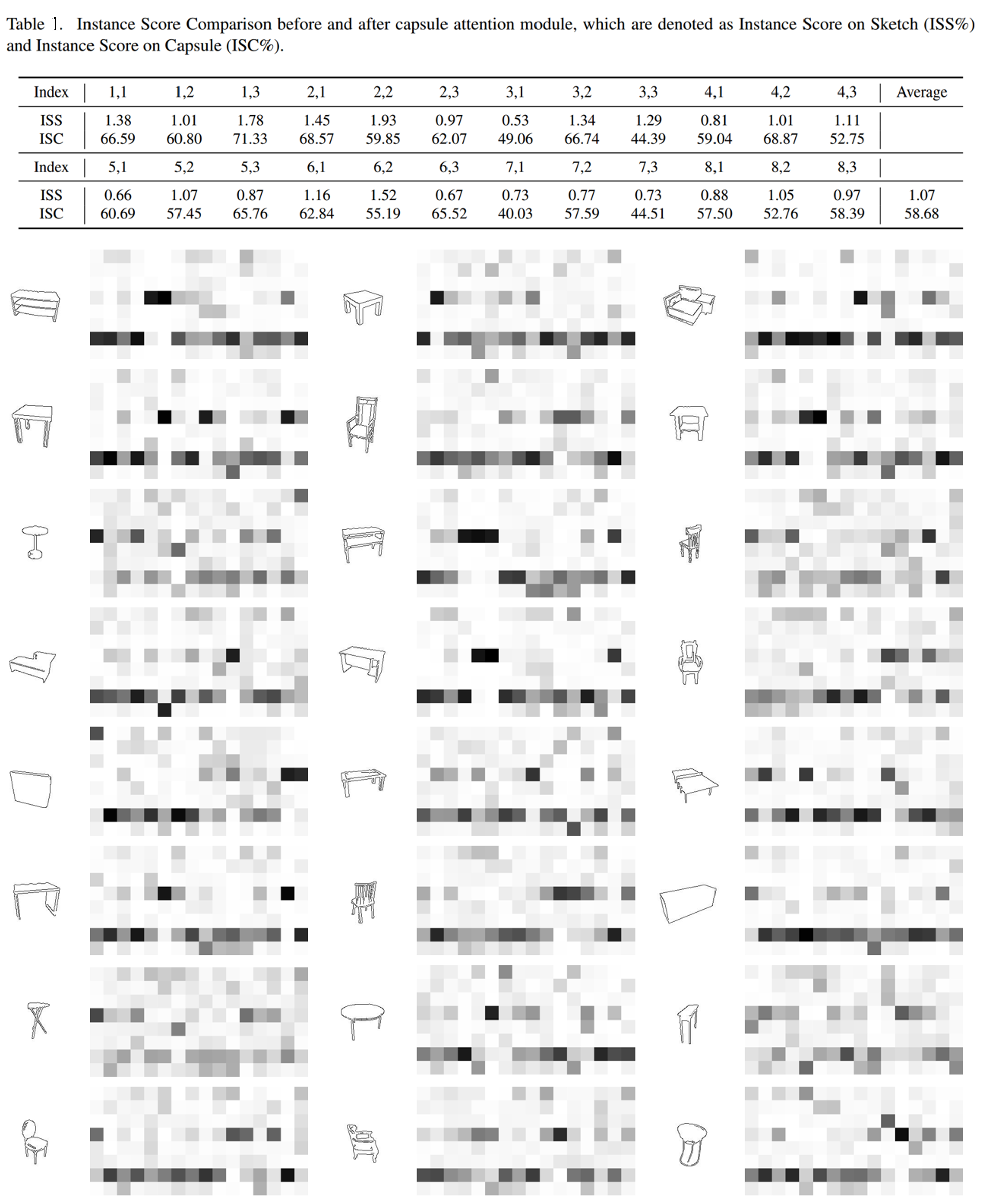}};

	\end{tikzpicture}
	
	\vspace{-3mm}
	\caption{\label{fig:attentions} Visualization of attention score maps from sketch to capsule. For attention score maps, the darker capsule components pay more attention with our attention-routing module in the final feature extraction. The lighter parts denote less attention. We report the Instance score in Table 2 to show the improvement of the attention.}
	\vspace{-4mm}
	\label{fig:attentionMaps}
\end{figure*}

All visualized sketch and attention score maps are reported in Table 2 and Figure~\ref{fig:attentionMaps}. Note that, across all 24 sketches, the average number of useful pixels representing the instance is only 1.07\% of the whole sketch. After our capsule-attention module, the attention weight of the instance improves to 58.68\% for better diffusion.

\section{More Point Cloud Generation Results}
\begin{figure*}[!htpb] 
	
	\begin{tikzpicture}[inner sep=0pt,outer sep=0pt]
		\centering
		\node[anchor=south west] (A) at (0in,0in)
		{\includegraphics[width=0.98\textwidth,clip=false]{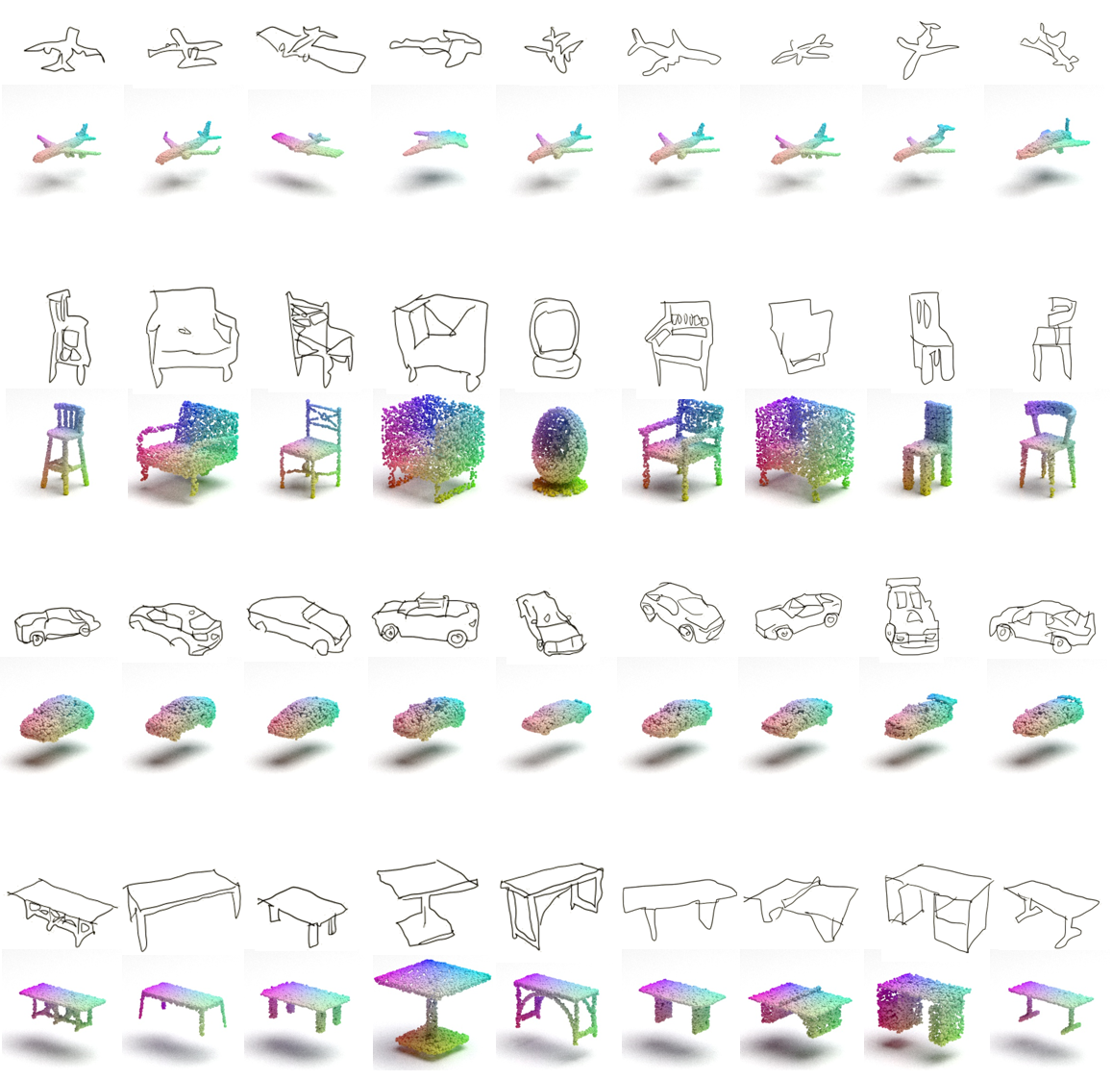}};

	\end{tikzpicture}
	
	\vspace{-3mm}
	\caption{\label{fig:gen1} Visualization of point cloud generations for $airplane,~chair,~car,~table$ category from free-hand sketches. Our model achieves good generalization ability for generating different shapes that are faithful to the condition provided by the respective sketch.}
	\vspace{-4mm}
\end{figure*}

We further test our STPD model for generating object shapes and categories from sketches. In this section, we focus only on shape and do not include appearance diffusion. Figure \ref{fig:gen1} shows our results. For each generation, we input the sketch to generate the 3D shape for all four categories ($airplane$, $chair$, $car$, and $table$) used in our experiments. Since there is no text description associated with 3D shape to support color diffusion, we manually assigned colors for better visualization. The numerous generation results demonstrate that our STPD has a good ability to generalize and generate different 3D shapes.

\begin{figure*}[!htpb] 
	
	\begin{tikzpicture}[inner sep=0pt,outer sep=0pt]
		\centering
		\node[anchor=south west] (A) at (0in,0in)
		{\includegraphics[width=0.98\textwidth,clip=false]{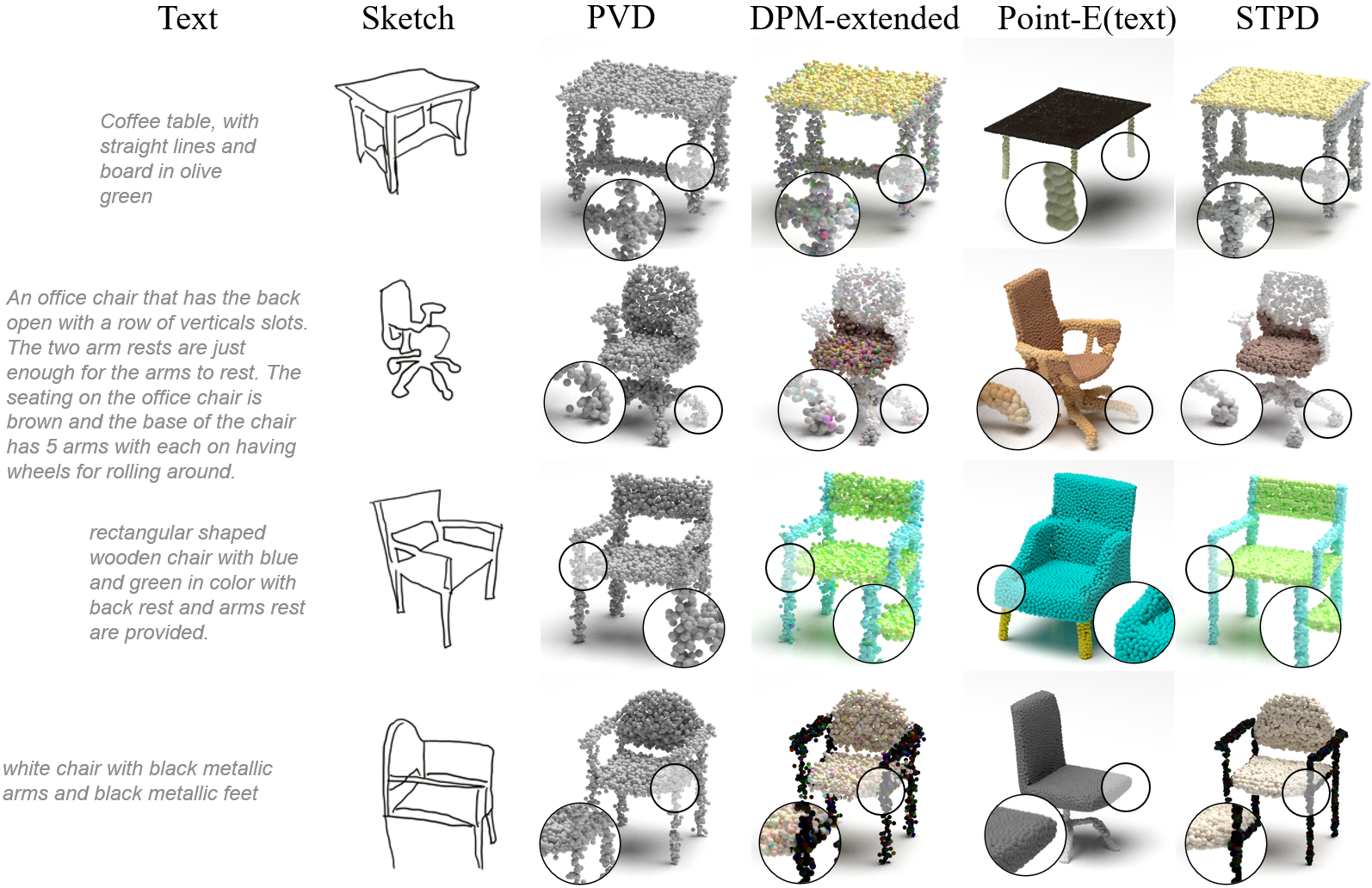}};

	\end{tikzpicture}

	\caption{\label{fig:comp} Visualization comparison Between PVD~\cite{zhou20213d}, DPM~\cite{diffusion1}, Point-E~\cite{nichol2022pointe}, and our STPD. We enlarge the local details to demonstrate that our STPD outperforms other baselines in terms of geometry and color generation. Our method produces faithful results from the input conditions.}
	\label{fig:attentionMaps2}
\end{figure*}

\section{Detailed Comparisons}
We provide a more detailed visualized comparison between PVD~\cite{zhou20213d}, DPM~\cite{diffusion1}, Point-E~\cite{nichol2022pointe}, and our STPD. PVD is trained without color. However, this approach has some drawbacks, such as not being able to capture the full range of colors present in a given image.

DPM has been extended to simultaneously capture color and geometry. However, the additional dimension for color may bias the shape and result in higher variance when simply applying the dimension extension on the sketch-text dataset.

Point-E is demonstrated here from a well-trained large dataset and performs great. However, there are two main problems with Point-E: 1) the shape generation does not strictly follow the conditions (e.g., conditions with black metallic arms but not generating the arms) and 2) the color is not accurately related to the shape parts as conditioned (e.g., showing black metallic feet but displaying gray feet).

In contrast, our method, STPD, shows faithful and fine results from the conditions.

\section{Additional Discussions}

In this section, we will discuss three features of our proposed method: (a) generating novel shapes, (b) the effects of slight changes in the sketch, and (c) conflicting sketch and text.

Our STPD method can learn the variance of parts from the training set. Given novel changes (such as back holes and hollow backrest sketch) that do not exist in the training shapes, the STPD learns these changes from the whole shapes and modifies the generated shapes to make the results faithful to the condition. However, we only trained the diffusion model on some categories of ShapeNet~\cite{chang2015shapenet}, which may limit its ability to deal with totally strange objects. It would be an interesting direction to train a more generalized model across large datasets using our extending methods, like the Stable-diffusion did in the 2D area.

We also tested some slight changes in the sketch to show that the results are faithful to the conditions. However, our vanilla STPD model is trained on a small dataset with sketch-text conditions, and we did not consider conflicts during training. Thus, we tested conflicting input sketches and text (such as a sketch with no armrest and text with an armrest). During the diffusion process, the model sometimes followed the sketch and sometimes the text. We believe that the well-trained BERT mask would randomly mask the conflict information in each modality.

\begin{figure*}[!htpb] 
	
	\begin{tikzpicture}[inner sep=0pt,outer sep=0pt]
		\centering
		\node[anchor=south west] (A) at (0in,0in)
		{\includegraphics[width=0.98\textwidth,clip=false]{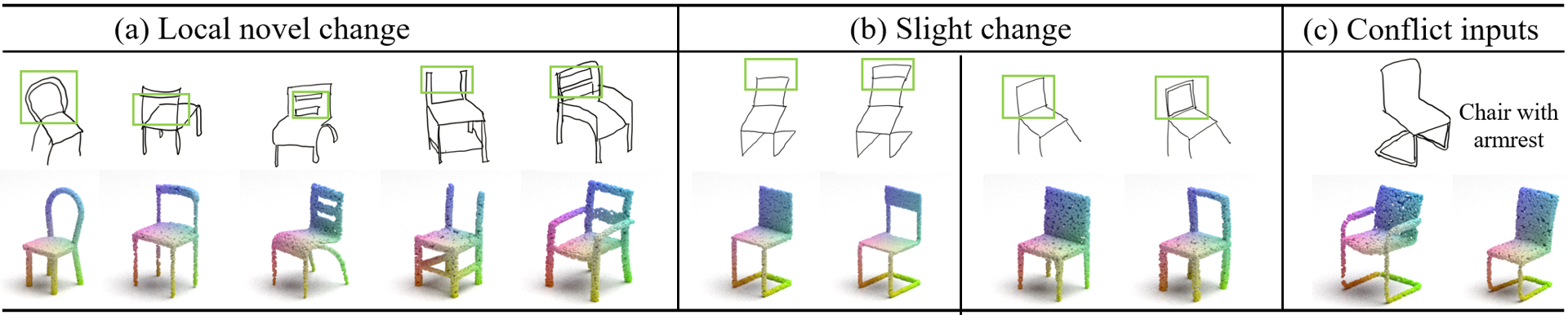}};

	\end{tikzpicture}
	
	\vspace{-1mm}
	\caption{\label{fig:gen2} Feature discussions. (a) generating novel shapes, and (b) effects of slight changes in the sketch, and (c) conflicting sketch \& text. The main change part of sketch is highlighted with a green box.}
	\vspace{-4mm}
\end{figure*}
{\small
\bibliographystyle{ieee_fullname}
\bibliography{egbib}
}

\end{document}